\documentclass[10pt, a4paper]{article}

\usepackage{multirow}
\usepackage{multicol}
\usepackage{cite}

\usepackage{tikz}
\usepackage{pgf}

\usepackage[]{lrec-coling2024} % this is the new style

\usepackage{listings}

\usepackage{natbib}
\usepackage{multibib}
\makeatletter
\def\@mb@citenamelist{cite,citep,citet,citealp,citealt,citepalias,citetalias}
\makeatother
\newcites{languageresource}{~}

\usepackage{graphicx}
\usepackage{tabularx}
\usepackage{amsmath}
\usepackage{soul}
\usepackage{booktabs}
\usepackage{xcolor}
\usepackage{hyperref}
 \definecolor{darkblue}{rgb}{0, 0, 0.5}
  \hypersetup{colorlinks=true, citecolor=darkblue, linkcolor=darkblue, urlcolor=darkblue}

\usepackage{xstring}

\usepackage{color}
\usepackage{float}
\usepackage{longtable}

\usepackage{subcaption}

\newcommand*\circled[1]{\raisebox{.5pt}{\textcircled{\raisebox{.5pt} {\fontsize{6pt}{6pt}\selectfont #1}}}}

\title{Generating Hard-Negative Out-of-Scope Data\\ with Chat{GPT} for Intent Classification}

\name{Zhijian Li,~ Stefan Larson,~ Kevin Leach} 

\address{Vanderbilt University, Nashville, TN, USA\\
\texttt{\{zhijian.li, stefan.larson, kevin.leach\} @ vanderbilt.edu}
}

\abstract{
Intent classifiers must be able to distinguish when a user's utterance does not belong to any supported intent to avoid producing incorrect and unrelated system responses.
Although out-of-scope (OOS) detection for intent classifiers has been studied, 
previous work has not yet studied changes in classifier performance against hard-negative out-of-scope utterances (i.e., inputs that share common features with in-scope data, but are actually out-of-scope).
We present an automated technique to generate hard-negative OOS data using ChatGPT.
We use our technique to build five new hard-negative OOS datasets, and evaluate each against three benchmark intent classifiers.
We show that classifiers struggle to correctly identify hard-negative OOS utterances more than general OOS utterances. 
Finally, we show that incorporating hard-negative OOS data for training improves model robustness when detecting hard-negative OOS data and general OOS data.
Our technique, datasets, and evaluation address an important void in the field, offering a straightforward and inexpensive way to collect hard-negative OOS data and improve intent classifiers' robustness.
 \\ \newline  
}

\begin{document}

\maketitleabstract

\section{Introduction}

Task-oriented dialog systems rely on robust intent classification models to produce appropriate responses based on the user utterances.
During deployment, the intent classifiers need to not only accurately classify the user utterances, but also must identify if user utterances do not belong to any supported intents.
Brittle intent classifiers that fail to reliably distinguish OOS (out-of-scope) utterances from the INS (in-scope) utterances ultimately lead to poor user experiences, wasted time and resources, and potential safety concerns.
Thus, it is imperative to develop techniques to ensure robustness against OOS utterances.

%As many businesses are  
% such as hotel booking \cite{li2019real}and an banking \cite{app112210995} 
%increasingly adopting conversational AI,
% as well as intent classifier's potential usage in controlling self driving vehicles \cite{weng2016conversational,okur2018conversational} and automated manufacturing machinery \cite{li2022tod4ir, baca2003dialog}, 
%it is imperative to ensure classifiers can reliably detect OOS utterances.
% ^ you aren't contributing to reliable detection, you are contributing a technique to improve data quality.  The conclusions are different and thus the introduction must reflect what you are actually solving 

\begin{figure}
    \centering\scalebox{0.55}{
    \includegraphics{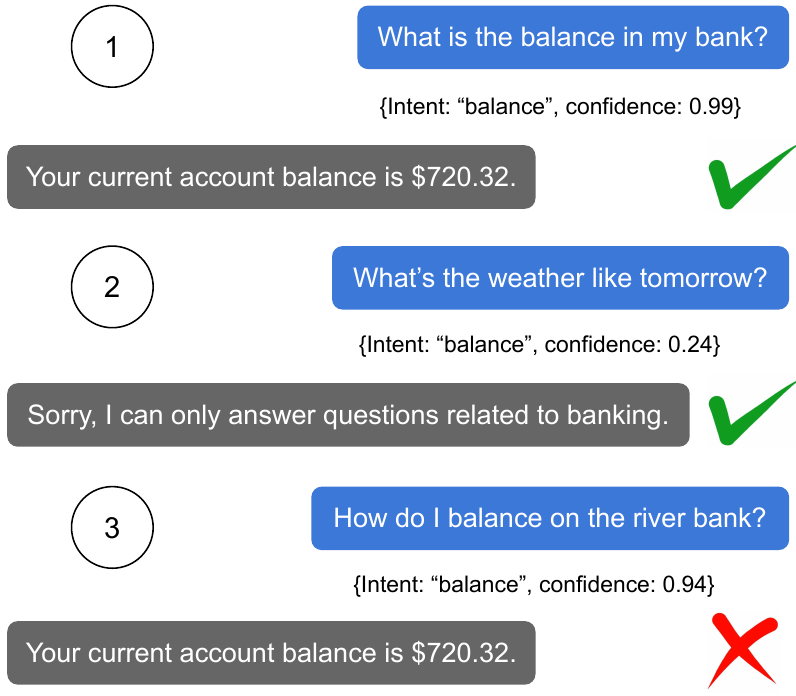}}
    \caption{Example exchanges between a user (blue,
right side) and a task-driven dialog system for personal
finance (grey, left side). The system correctly identifies the user’s utterance as in-scope in \circled{1}, and correctly identifies the user's utterance as out-of-scope and gives a valid response in \circled{2}. In \circled{3}, the system incorrectly identifies the hard-negative OOS user utterance as in-scope and provides an incorrect response.}
    \label{conv_demo}
\end{figure}

When developing intent classifiers, developers typically begin by collecting and labelling a large amount of INS training data often acquired through crowd-sourcing \cite{kang-etal-2018-data, larson-2022-survey}.
Collecting OOS datasets to improve the classifier's OOS detection capability is not a common practice~\cite{larson-etal-2019-evaluation}.
Currently, there is a dearth of large public OOS datasets, and the difficulty posed to intent classifiers has not been rigorously examined.
General OOS datasets typically contain mostly samples that exhibit minimal similarity with the INS samples.
While many models can distinguish such OOS samples and the INS samples during testing, using such OOS samples to evaluate models' OOS detection capabilities can produce misleading results.
During deployment, the model may encounter OOS utterances that closely resemble the INS utterances but convey entirely different meanings.
Due to the high similarity that those \textit{hard-negative out-of-scope} utterances share with the INS data, the models are more susceptible to misclassifying them into a supported intent with high confidence.
% When those utterances do not belong to any of the supported intents, the models are more susceptible to misclassifying them into a supported intent.
Therefore, collecting hard-negative OOS data is pivotal to ensure that the intent classifiers can reliably distinguish all OOS utterances, regardless of their resemblance to the INS samples.

Figure~\ref{conv_demo} shows example utterances and responses of a task-oriented dialog system driven by an intent classifier trained only on utterances for personal finance.
In the first user-system exchange, the system correctly categorizes the user utterance as an in-scope \texttt{balance} utterance with appropriately high confidence.
In the second exchange, the user provides a general OOS utterance, and the system successfully identifies that the user utterance cannot be understood because the confidence score for all the intents are low.
In the third exchange, the user presents a hard-negative OOS utterance.
Since the utterance includes the words ``balance'' and ``bank'', and appears similar to the in-scope training data for the \texttt{balance} intent, the system incorrectly classifies the utterance as in-scope for \texttt{balance} and produces an incorrect response.

Obtaining OOS data that sufficiently challenges the intent classifiers is difficult.
Often, this is done through crowd-sourcing with platforms like Amazon Mechanical Turk (e.g., \citet{larson-etal-2019-evaluation}).
However, this approach is costly and time-consuming, and it requires careful verification to make sure that the samples are indeed out-of-scope and challenging.
Furthermore, data collection via crowd-sourcing introduces quality control problems as the the collected data are often error prone \cite{shah2016no, larson-etal-2020-inconsistencies}.
% Additionally, there are currently no known techniques to aid crowd-source workers in identifying OOS utterances. 
In datasets containing a large number of intents, verifying each utterance to be irrelevant from all the intents poses significant difficulty for human crowd-sourcing workers.
For example, the Clinc-150 dataset \cite{larson-etal-2019-evaluation} encompasses 150 intents;
relying on the crowd-workers to verify all sentences to be OOS may be challenging and could lead to erroneous results. 

A cost-effective alternative is to generate hard-negative OOS data using Large Language Models such as ChatGPT. \citet{larson-etal-2019-evaluation} paid crowd workers \$0.20 US to produce three paraphrases of an utterance, and current best practices involve payment that extrapolates to a fair wage \cite{kummerfeld-2021-quantifying}.
%Considering the challenge of producing high quality hard-negative OOS sentences, the cost this task using crowd sourcing would likely be higher.
In comparison, the GPT-3.5 turbo API costs \$0.0015 per 1K tokens in the prompt and \$0.003 per 1K tokens for the output, a substantial potential savings.

%which is much cheaper than the crowd-sourcing approach.

In this paper, we aim to investigate the following research questions: 

\begin{enumerate}
\itemsep-2pt
    \item{Can ChatGPT generate OOS utterances that do not fall into any of the system-supported INS intents?}
    \item{Do the generated hard-negative OOS utterances lead to high-confidence predictions from intent classifiers?}
    \item{Can the generated hard-negative OOS utterances be used in training to improve the intent classifiers' OOS detection ability and decrease the models' confidence on the supported intents when encountering OOS utterances?}
\end{enumerate}

To answer these questions, we select five large public datasets and introduce a method to generate 3,732 hard-negative OOS utterances --- that is, utterances intended to closely resemble the in-scope data for a given intent. 

Our method works by analyzing important words that are likely to have the biggest influence on the intent classifiers' predictions for each intent from the INS training data.
Then, the method prompts ChatGPT to generate OOS utterances that include specific important words for each intent.
Since the generated OOS utterances contain the important words from supported intents, they are more likely to confuse the model and produce high confidence. 
In doing so, we can increase the rigor and challenge of a given dataset by partially automating the creation of hard negative inputs. 

We evaluate the performance of three benchmark transformer models for intent classification on our generated datasets. 
The hard-negative OOS utterances produced using our method consistently resulted in high confidence (but incorrect) predictions across all five datasets.
Notably, intent classifiers struggled to distinguish the hard-negative OOS utterances from the INS utterances. 
In particular for Clinc-150 and Banking77, model confidence scores are substantially higher for our generated hard-negative OOS datasets compared to the general OOS dataset.
%After incorporating the generated hard-negative OOS data in training, the models' confidence for the supported intents decreased substantially. 
We see various improvements in model performance when incorporating hard-negative OOS data in training.
For instance, for Banking77 evaluated with BERT, the AUROC for hard-negative OOS and INS improved to 0.996 and the AUROC for general OOS and INS improved to 0.989 when the model is trained on both INS and hard-negative OOS data.
Our study shows that more attention must be given to curating hard-negative OOS datasets to enhance model robustness in deployment.

\section{Related Work}

In this section, we discuss prior work related to data collection for intent classification tasks and ChatGPT's data generation capabilities.

\subsection{Hard-Negative Data}

Several studies have highlighted the significance of using hard-negative samples during training to improve model robustness.
%In \citet{zhan2021optimizing}'s study, they utilized hard-negative examples during training to aid retrieval models to better discriminate between relevant and irrelevant documents.
For instance, \citet{zhan2021optimizing} and \citet{nguyen-etal-2023-passage} used hard-negative examples during training to aid retrieval models to better discriminate between relevant and irrelevant documents.
Hard-negatives can also facilitate contrastive learning for image classification (e.g., \citet{kalantidis2020hard}) and image retrieval (e.g., \citet{siamese-image-matching, rs10101552-remote-sensing-hard-negatives}). However, the use of hard-negatives for intent classification remains understudied. In this paper, we use our generated hard-negative OOS datasets to improve model robustness for intent classification.

% stefan started this subsection
\subsection{Data Collection for Intent Classification}

\textbf{General Data Collection.} Prior strategies for data collection for constructing training and evaluation data for intent classifiers include the use of crowd-sourcing to generate queries by either (1) paraphrasing input query prompts or (2) responding to scenarios with queries.
This prior work includes \citet{coucke2018snips, larson-etal-2019-evaluation, gupta-etal-2018-semantic-parsing, hwu64-2019, kang-etal-2018-data}.
Recently, prior work has investigated using large language models (LLMs) to generate this type of training data.
This work includes \citet{rosenbaum-etal-2022-linguist}, who used AlexaTM, and \citet{sahu-etal-2022-data}, who used GPT-3, and \citet{cegin-chatgpt-crowdsourcing-2023}, who used ChatGPT to generate paraphrases.

%\subsection{Out-of-Scope and Challenging Data Collection}
\noindent\textbf{Out-of-Scope and Challenging Data Collection.}
Most prior work on data collection for intent classification does not consider the production of OOS queries for evaluating a model's ability to distinguish between in- and out-of-scope inputs.
Exceptions to this include \citet{larson-etal-2019-evaluation}, who's Clinc-150 dataset includes out-of-scope utterances that were generated via crowd-sourcing.
Other work in this space includes \citet{larson-leach-2022-redwood}, whose out-of-scope data is constructed by sampling from other datasets, and \citet{zhang2021pretrained}, who constructed ``in-domain, out-of-scope'' splits of Clinc-150 and Banking77 where training data includes a set of intents $I_{in\text{-}scope}$ from the original dataset $I$, but evaluation data includes a set of intents $I_{out\text{-}of\text{-}scope}$ that are in the same domain\footnote{~Here, a domain is a semantically meaningful group of intents like ``banking'', ``travel'', etc.} as $I_{in\text{-}scope}$ but $I_{in\text{-}scope} \cap I_{out\text{-}of\text{-}scope} = \emptyset$. 

Similarly, \citet{khosla-gangadharaiah-2022-benchmarking} created challenging datasets for testing model robustness to ``covariate shifted'' data.
In that work, covariate shifted data means data that was generated from a different distribution with respect to some other original data distribution, but where both data distributions generate data for the same intent category.
Considering the \texttt{weather} intent from Clinc-150, covariate shifted data includes data that was originally generated for the Snips or HWU64 datasets (i.e., the \texttt{get\_weather} intent from Snips, and HWU64's \texttt{weather\_query} intent).

% \fullspace Both \citet{khosla-gangadharaiah-2022-benchmarking} and \citet{zhang2021pretrained} create challenging data using already-generated data from existing corpora.
Other work that focuses on directly generating new challenge data includes techniques from \citet{larson-etal-2019-outlier}, which prompted crowd workers to paraphrase only unique queries from a dataset, and \citet{larson2020iterative}, who prompted crowd workers to paraphrase seed phrases but constrained the crowd workers from using certain keywords that were found to be indicative of certain intents.
The techniques from both of these works can be seen as ways to generate covariate shifted in-scope data.

\subsection{Adversarial Examples}
Our work on generating hard-negative data has similarities to prior work on generating adversarial examples.
Motivated by adversarial example generation work in computer vision that applies ``imperceptible'' perturbations to images to provoke incorrect model classification (e.g., \citet{adversarial-examples-goodfellow-2015}), prior work in natural language processing on adversarial example generation has revolved around perturbing texts using character-level alterations (e.g.,~\citet{ebrahimi-etal-2018-hotflip, gao-2018-black-box-adversarial}) and word synonym and phrase replacement (e.g.,~\citet{alzantot-etal-2018-generating, ren-etal-2019-generating, garg-ramakrishnan-2020-bae}).
Similar work in dialog data has been done by \citet{peng-etal-2021-raddle, liu-etal-2021-robustness, sengupta-etal-2021-robustness}, where utterances are modified (e.g., by introducing typos, word synonyms, ASR errors, etc.) in order to test the robustness of models.
The method we present in this paper can be seen as way to generate adversarial examples to test the robustness of intent classification models, but differs in that we do not apply perturbations to existing samples.

% \subsection{Hard-Negative OOS Data Collection: Crowd-sourcing vs ChatGPT}
% Collecting OOS data through crowd-sourcing is difficult, and there is no current work on creating hard-negative OOS datasets for intent classification. 
% In \citet{larson-etal-2019-evaluation}'s study, when they prompted the crowd-sourcing workers to produce OOS utterances, only 69\% of the utterances were out-of-scope. 
% Verifying that all the utterances are correctly labeled required expert knowledge over all the intents in the dataset. It would be reasonable to assume that trying to generate OOS data that also look similar to the INS data with crowd-source workers would be more challenging than collecting general OOS data, and the quality of the data would be hard to ensure.

\begin{figure*}[!ht] 
  \centering\scalebox{1}{
  \includegraphics[width=\linewidth]{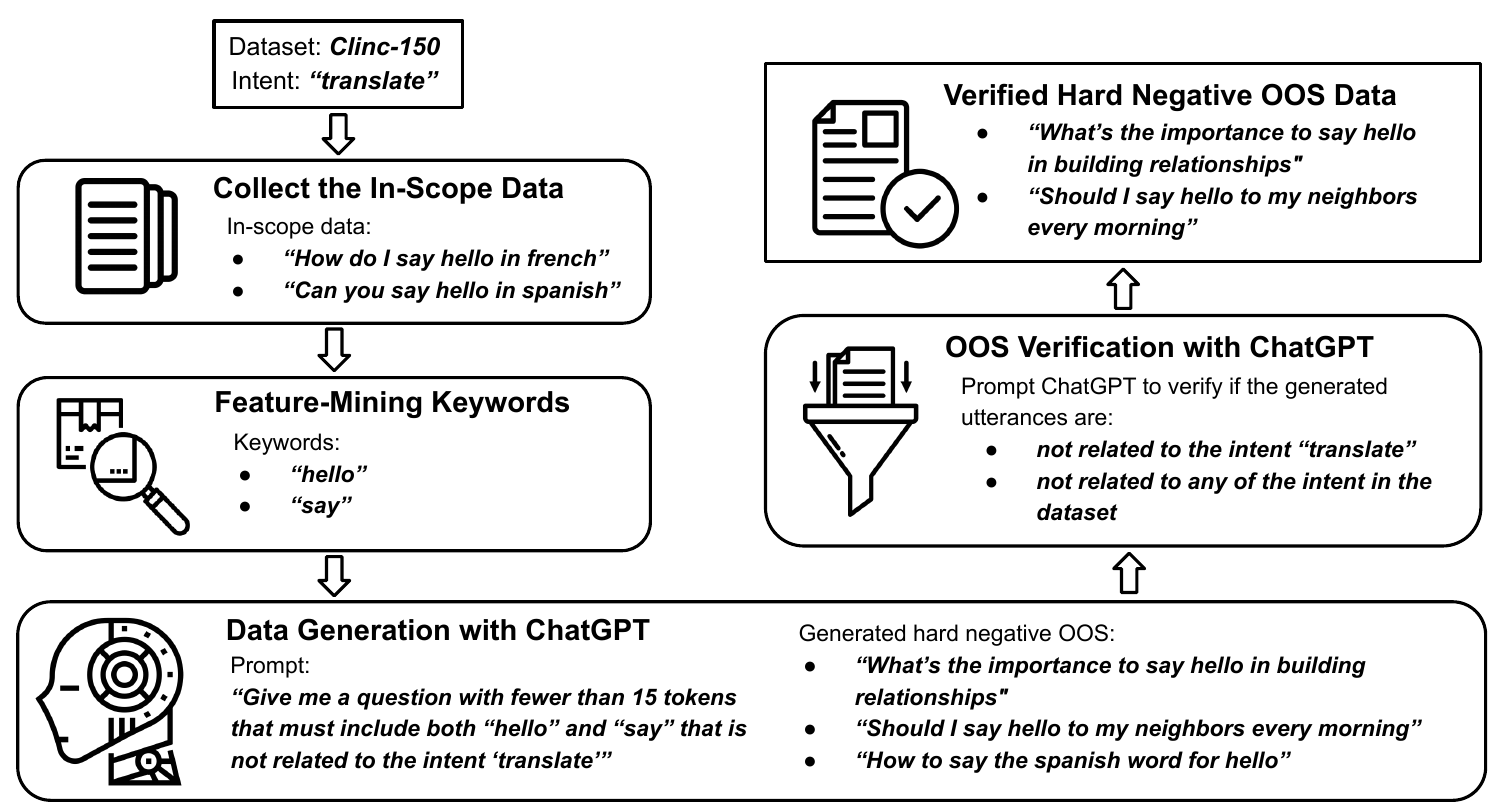}}
  \caption{An overview of the hard-negative OOS generation process, including examples. The third generated utterance is filtered out during the two-step OOS verification.}
  \label{overview}
\end{figure*}

\subsection{Data Collection and Annotation with ChatGPT}

In a recent study,~\citet{cegin-chatgpt-crowdsourcing-2023} showed that large language models such as ChatGPT can generate more lexically and syntactically diverse INS data by paraphrasing existing corpora. 
\citet{cegin-chatgpt-crowdsourcing-2023} also showed that ChatGPT can follow prompted restrictions and used ChatGPT in-lieu of crowd-sourcing to generate ``taboo'' paraphrases in the manner of \citet{larson2020iterative}, where certain words are avoided in paraphrases in order to promote diversity.
%ChatGPT can also reliably follow instructions in a prompt to avoid the usage to taboo words~\cite{larson2020iterative} in the generated sentence to improve sentence diversity. 
However,~\citet{cegin-chatgpt-crowdsourcing-2023} also observed several issues with open-sourced models such as Falcon-40b: duplicated outputs, erroneous outputs, and lack of instruction following. 
% Cegin et al. (2023) found that data produced by Falcon-40b was less diverse and lower quality than ChatGPT.
Another study showed ChatGPT's performance varies for sentiment analysis on tweets depending on the topics ~\cite{zhu2023can}. For tasks such as classifying the political affiliation of Twitter users, ChatGPT outperforms human crowd-source workers~\cite{tornberg2023chatgpt}. 
ChatGPT's performance for text-annotation tasks has also been found to exceed that of crowd workers (e.g.,~\citet{gilardi2023chatgpt, crowdsource-gpt4-he-2024}). 
\citet{sahu-etal-2022-data} prompted GPT-3 to generate labeled training data, and the generated data significantly improves the intent classifiers when the intents are distinct. Using ChatGPT to rephrase sentences, models trained with AugGPT outperform state-of-the-art text data augmentation methods to generate data for scarce intents in a few-shot learning setting~\cite{AugGPT-2023}.

% ...more ... \citet{rosenbaum-etal-2022-linguist, data-aug-chen-2022, yoo-etal-2021-gpt3mix-leveraging, fabricator-llm-2023}

\section{Methods}

We introduce an automated method for generating hard-negative OOS utterances using ChatGPT's API.
In this study, we generate 3,732 hard-negative OOS queries for five benchmark datasets.
Our objective is to generate hard-negative OOS utterances by producing utterances that are likely to contain words that heavily influence the intent classifiers' predictions.

The hard-negative OOS datasets are generated following these steps, which we discuss throughout this section.
Example utterances produced with our approach are shown in Figure~\ref{overview}.

\vspace{-1mm}
\begin{enumerate} 
    \item{Use feature-mining to select $n$ keywords by analyzing the most frequently appearing words for each intent for every selected dataset}
    \vspace{-2mm}
    \item{Select a combination of $m$ keywords $k_1$ ... $k_m$ from the $n$ keywords for an intent $i$}
    \vspace{-2mm}
    \item{Show ChatGPT the name of the intent $i$ and five in-scope utterances from $i$}
    \vspace{-2mm}
    \item{Prompt ChatGPT to generate $x$ questions (i.e., utterances) that must contain $k_1$ ... $k_m$, and is not related to the intent $i$}
    \vspace{-2mm}
    \item{Prompt ChatGPT to verify that each of the $x$ generated questions is not related to the intent $i$}
    \vspace{-2mm}
    \item{Prompt ChatGPT to verify that each of the $x$ generated questions is not related to any of the intents in the entire dataset}
\end{enumerate}

In this study, we select $n$ to be $5$, $m$ to be $2$, and $x$ to be $4$. When the prompts include only one keyword, we noticed that the generated data shares less similarity to the INS data compared to when $m$ = $2$. When prompts include three or more keywords, ChatGPT frequently struggles to include all the keywords or produce an OOS utterance. Larger $n$ and $x$ can be selected to generate more hard-negative OOS data.

\subsection{Feature-Mining Keywords}

To ensure that the generated OOS queries resemble INS queries for each intent, we need to identify important keywords that are likely to influence the intent classifiers' prediction.
Among our five selected datasets, we collect the most frequently occurring words in the INS training samples from each intent.
We lemmatize all words using NLTK's WordNetLemmatizer \cite{bird-loper-2004-nltk} to prevent getting multiple versions of the same word.
In addition, we discard the stop words and tokens that contain less than three characters.
We have also explored alternative methods to identify keywords.
For instance, using the Python ELI5 package,\footnote{~\url{https://github.com/eli5-org/eli5}} we determined the words with the highest weights after training an SVM classification model on each of the datasets, as was done in \citet{larson2020iterative}.
For larger datasets such as Clinc-150 and HWU64, training multi-class SVM on a substantial number of intents is time consuming. We utilized the LinearSVC from Scikit-Learn \cite{pedregosa2011scikit} which is implemented as One-vs-All, resulting in a model requiring $n$ classifiers for $n$ intents. 
After removing stopwords and lemmatizing the tokens, we are not able to obtain at least five keywords for many intents. For example 65 of 150 intents from Clinc-150 produced less than 5 keywords using the ELI5 method. 
Therefore, we use the original token frequency-based approach for keywords collection.

%\subsection{ChatGPT Data Generation}
\subsection{Data Generation with ChatGPT}

Recall that we prompt ChatGPT to produce hard-negative OOS samples. 
We use the GPT 3.5-turbo model through the chat completion API. (All experiments using ChatGPT were done June-August 2023.) For interaction with the API, we make use of three distinct ``roles'': (1) The ``system'' role, which allows the developers to guide ChatGPT's behavior throughout the conversation. 
(2) The ``assistance'' role, which grants ChatGPT's API the access to previous conversations, enabling ChatGPT to recall previously generated hard-negative OOS utterances to avoid generating duplicates.
ChatGPT can also retain the intent information, removing the need for us to repeatedly inform ChatGPT the intent in every prompt. 
(3) The ``user'' role, which lets the developers to prompt questions for ChatGPT's API to answer.

When generating hard-negative OOS utterances, we first set the role to ``system'' and guide ChatGPT to answer with only the hard-negative sentence.
This approach prevents the API from padding the responses with unnecessary tokens like ``Sure, I'd be happy to help.'' 
Next, we set the role to ``user'' and show ChatGPT the name of each intent and five INS samples for that intent.
This enables ChatGPT to understand the semantics of the intent, especially when the intent name alone does not provide sufficient context.
Then, we use the ``assistance'' role to record the dialog between the developer and ChatGPT. 

Subsequently, we prompt ChatGPT to generate an utterance that must be unrelated to the intent (i.e., OOS) and must contain a combination of two keywords 
%$k_1$ and $k_2$ 
collected during feature-mining.\footnote{~Preliminary experiments revealed that ChatGPT will often output utterances with mostly the same tokens if prompted to generate multiple utterances at once.}
This process allows us to guide ChatGPT to produce an utterance that contains commonly used words associated with a given intent but that is not related to that intent --- that is, a hardnegative OOS sample. 
Next, we discuss how we validate each such generated utterance is in fact OOS. 
% XXX I don't understand the next sentence
%In some cases, a combination of two keywords leads heavily in the direction of the intent, resulting in difficulties including them in an OOS query.
%In such cases, ChatGPT often produces utterances that are either INS or contain only one of the required words. 
% stefan: fixme the above two sentences might distract too much and confuse the reviewer. same for the footnote

%\subsection{ChatGPT OOS Verification}
\subsection{OOS Verification with ChatGPT}

To ensure that the hard-negative OOS data does not contain any INS samples, we use a two-step verification method using ChatGPT.
After ChatGPT generates every hard-negative OOS utterance, we immediately prompt ChatGPT to assess whether each utterance belongs to the intent that the utterance should \emph{not} relate to.
For example, after prompting ChatGPT to generate a question containing ``hello'' and ``french'' that is not related to the \texttt{translate} intent, we subsequently prompt ChatGPT to determine whether the generated utterance is related to ``translate''.
If ChatGPT determines that the utterance is related to ``translate'', then it is discarded.
In the second step of verification, the remaining utterances are then checked to be OOS with respect to the entire dataset.
To implement this step, we provide ChatGPT with the name of all INS intent categories using the ``system'' role, and prompt ChatGPT to verify if each utterance is OOS with the `user' role.

We note that ChatGPT occasionally mislabels a small portion of the utterances during this two-step verification process.
However, since our goal is to generate hard negative OOS data, discarding such inadvertently mislabeled data is not critical. 
%Recall that our goal is to generate OOS data, the inadvertent discarding of valid hard-negative OOS by ChatGPT is not an paramount concern as we still have a sufficient number of utterances in our datasets to carry out the evaluation and conclusion.
Finally, we manually check the hard-negative OOS datasets collected after the two-step verification stage to ensure the label accuracy.
For utterances that do not clearly fall into the INS or OOS categories, we compare the utterance with the INS samples and discuss amongst the research team to conclude the correct label. 
We discard the utterances if no consensus is reached regarding label opinions.

\section{Evaluation}

%In order to assess the level of difficulty in distinguishing the generated hard-negative OOS datasets from the general OOS datasets by intent classifiers, we evaluated the performance of various classifier models using the hard-negative OOS data.
We design experiments to assess the level of difficulty that intent classifiers have when discerning between in-scope (INS) and hard-negative out-of-scope (OOS) data.
As a baseline, we also measure models' abilities to discern INS data from ``general'' (i.e., not hard-negative) OOS data.
Additionally, we evaluate the improvements in models' OOS detection abilities after including hard-negative OOS and general OOS data in training.
We hypothesize that the hard negative OOS data we generate with our approach will lead to performance decreases because the model will confuse these samples as being INS.

%\subsection{In-Scope Datasets}

\subsection{Data}

\paragraph{In-Scope Data.}
We use INS data from \textbf{Clinc-150}~\cite{larson-etal-2019-evaluation}, \textbf{Banking77}~\cite{casanueva-etal-2020-efficient}, \textbf{ATIS}~\cite{hemphill-etal-1990-atis,hirschman-1992-multi,hirschman-etal-1993-multi,dahl-etal-1994-expanding}, \textbf{Snips}~\cite{coucke2018snips}, and \textbf{HWU64}~\cite{hwu64-2019}.
Different from the other four datasets, HWU64 consists of two intents \texttt{General\_Quirky} and \texttt{QA\_Factoid} that span a wide spectrum of semantics. 
We excluded these two ``catch-all'' intents from HWU64 for training and testing to allow for sufficient samples that are considered OOS.

\paragraph{Hard-Negative Out-of-Scope Data.}
We use our method to generate 3,732 hard-negative out-of-scope samples targeting intents from the five datasets listed above. Section~\ref{hard_negative_count} discusses the generated data in detail.

\paragraph{General Out-of-Scope Data.} Instead of hard-negative OOS data, the baseline approach uses ``general'' OOS data.
We use the Clinc-150 companion OOS data as this general OOS data, which consists of 1,000 test utterances.
We then filtered out any utterances that belonged to any of the in-scope intents from the other four datasets.
%To compare the intent classification models' performance on the hard-negative OOS dataset against both the INS data and the traditional OOS data, we must have an OOS dataset for each of the benchmark dataset. Among the five benchmark datasets, only Clinc-150 provides OOS queries. We collected the all the 1000 testing OOS queries from Clinc-150 and manually filtered out the INS utterances for each of the other four datasets for the evaluation process. For utterances that are not obvious to be INS or OOS, the appropriate label is discussed amongst the group.
\paragraph{Data Splits.}
We split the INS data into training and testing.
Each selected model is trained on the training data
and evaluated with the INS testing data, the generated hard-negative OOS data, the general OOS data from Clinc-150.
To examine whether hard-negative OOS data
can be used during training to improve the models’
OOS detection capability, we separated both hard-negative OOS data and general OOS data into
80\% training and 20\% testing splits and compared
the models’ confidence for the OOS datasets
when we included hard-negative OOS, general
OOS, and both OOS corpora in training.

\subsection{Intent Classification Models}

We use the following intent classifiers in our experiments:
\textbf{BERT}, a neural network that uses a transformer model to capture the context of each word and fine-tuned on the training data for NLP tasks \cite{devlin-etal-2019-bert}. In this study, we use bert-base-uncased.
\textbf{RoBERTa}, a variant of BERT that provides better contextual representation of text \cite{liu2019roberta}.
\textbf{DistilBERT}, a lightweight variant of BERT that uses fewer parameters than BERT and processes texts faster \cite{sanh2019distilbert}.
We used the Hugging Face implementations of these models \cite{wolf-etal-2020-transformers}.
% todo fixme: ideally, we can cite papers that use these models to show that they are not random models

Common approaches to dealing with OOS inputs involve the use of confidence scores to differentiate between INS and OOS inputs.
Normally, a desirable model is one that assigns higher confidence to INS inputs, and lower confidence to OOS inputs.
We use two functions to produce the confidence scores when evaluating our hard-negative OOS datasets:
(1) \textbf{Softmax}, where we use the highest softmax confidence score for each prediction~\cite{hendrycks2016baseline}.
(2) \textbf{Energy}, where we compute the energy score~\cite{liu2020energy} for each prediction using $T=1$.\footnote{~We considered alternative values for $T$ and the results are similar for different $T$.}
These two methods are commonly used in prior work on out-of-distribution detection (e.g., \citet{larson-2022-rvlcdip-ood}).

\subsection{Metrics}
We consider several performance metrics to evaluate the quality of our hard-negative OOS datasets, assigning the INS predictions as the positive class and OOS predictions as the negative class:
(1) \textbf{AUROC}, the Area Under the Receiver Operating Characteristic curve can be interpreted as measuring the overlap between two distributions and is commonly used for benchmarking OOD performances~\cite{fort2021exploring}. An AUROC score of 0.5 means the model is unable to produce confidence scores that distinguish between INS and OOS inputs, and an AUROC closer to 1.0 indicates better OOS detection capability. 
(2) \textbf{AUPR}, the Area Under the Precision and Recall Curve. A higher AUPR indicates a more robust model.
(3) \textbf{FPR95}, the false positive rate at 95\% recall. A lower FPR95 indicates a more robust model.
(4) \textbf{F1 score with confidence thresholds}, the F1 score for a range of confidence thresholds from 0.5 to 0.95 for softmax confidence scores. Predictions with higher confidence score than the confidence threshold are considered as positive predictions. Higher F1 captures the classifier's ability to distinguish INS and OOS data at different confidence thresholds since most intent classifiers in deployment utilize a confidence threshold to determine whether an utterance is not understood.

\begin{table*}
    \centering\scalebox{0.9}{
    \begin{tabular}{llll}
        \toprule
        \textbf{Dataset} & \textbf{Intent} & \textbf{Keywords} & \textbf{Hard-Negative OOS}\\
        \midrule
        Clinc-150 & find\_phone & find, locate & \emph{how do i \underline{find} and \underline{locate} a lost pet} \\
        Clinc-150 & change\_language & french, english & \emph{are \underline{french} and \underline{english} official languages in canada} \\
        Clinc-150 & distance & take, long & \emph{does it \underline{take} \underline{long} to find a seat on the bus} \\
        Banking77 & exchange\_rate & foreign, know & \emph{do you \underline{know} any \underline{foreign} exchange officers nearby} \\
        Banking77 & age\_limit & age, children & \emph{what are \underline{age}-appropriate money lessons for \underline{children}} \\
        Banking77 & card\_arrival & track, sent & \emph{can i \underline{track} my \underline{sent} documents} \\
        HWU64 & iot\_hue\_lightchange & color, change & \emph{can temperature \underline{change} the \underline{color} of a blue flame} \\
        HWU64 & alarm\_set & wake, tomorrow & \emph{what are some tips to \underline{wake} up refreshed \underline{tomorrow}} \\
        HWU64 & iot\_coffee & make, machine & \emph{how do i  \underline{make} a homemade washing \underline{machine}} \\
        ATIS & flight\_time & time, open & \emph{what \underline{time} does the golden gate bridge \underline{open}} \\
        ATIS & aircraft & type, used & \emph{what \underline{type} of currency is \underline{used} in boston} \\
        Snips & BookRestaurant & book, reservation & \emph{can i \underline{book} a \underline{reservation} for the flight} \\
        Snips & GetWeather & like, going & \emph{what's the vibe \underline{like} \underline{going} there} \\
        \bottomrule
    \end{tabular}}
    \caption{Example hard-negative OOS data generated by ChatGPT across different intents and datasets.}
    \label{example_hard_negative}
\end{table*}

\begin{table}
    \centering\scalebox{0.9}{
    \begin{tabular}{lrrrr}
        \toprule
        \textbf{Dataset} & \textbf{Total} & \textbf{Step 1} & \textbf{Step 2} & \textbf{Final} \\
        \midrule
        Clinc-150 & 6,000 & 4,442 & 2,278 & 2,266\\
        Banking77 & 1,440 & 1,169 & 742 & 734\\
        ATIS & 640 & 458 & 226 & 220\\
        Snips & 280 & 200 & 95 & 90\\
        HWU64 & 2,720 & 1,903 & 438 & 422\\
        \midrule
        Overall & 11,080 & 8,172 & 3,779 & 3,732\\
        \bottomrule
    \end{tabular}}
    \caption{Results of prompting ChatGPT to generate hard-negative OOS utterances for each dataset. ``Step 1'' and ``Step 2'' show how many utterances remained valid after the ChatGPT OOS Verification. 
    The total count of valid hard-negative OOS utterances is displayed in the ``Final'' column.}
    \label{hard_negative_count_table}
\end{table}

\section{Results}
This section discusses the results of our data generation and experiments: Section~\ref{hard_negative_count} discusses the results of generating hard-negative OOS with our method.
Section~\ref{baseline_evaluation} discusses results of experiments determining the effectiveness of hard-negative OOS data vis-\`{a}-vis general OOS data.
Section~\ref{sec:results-training} discusses results of experiments on determining the effectiveness of training with hard-negative OOS data.

\subsection{Generation Results} \label{hard_negative_count}
In this study, we prompted ChatGPT to generate a total of 11,080 hard-negative OOS utterances from five different datasets.
Roughly 74\% (8,172 samples) of the generated hard-negative passed the first round of OOS verification.
Then, 3,779 (34\%) remained to be valid after the second round of OOS verification where each utterance is prompted to ChatGPT to determine whether it is related to any known in-scope intent of the dataset.
Finally, when we manually examined the hard-negative OOS datasets after the two-step verification method, we found that 47 (1.2\% of the hard-negative OOS after the two-step verification) are INS and mislabelled by ChatGPT, resulting in 3,732 valid hard-negative OOS samples.
Examples of generated hard-negative data are shown in Table~\ref{example_hard_negative}.
Precise counts of verified hard-negative OOS data for each dataset is shown in Table~\ref{hard_negative_count_table}.

\begin{table*}
    \centering\scalebox{0.9}{
    \begin{tabularx}{\textwidth}{lXXXXXX}
        \toprule
        \multirow{2}{*}{\textbf{Dataset}} & \multicolumn{2}{c}{\textbf{AUROC}} & \multicolumn{2}{c}{\textbf{AUPR}} & \multicolumn{2}{c}{\textbf{FPR95}}\\
        \cline{2-7}
        & \textbf{General OOS} & \textbf{HN OOS} & \textbf{General OOS} & \textbf{HN OOS} & \textbf{General OOS} & \textbf{HN OOS} \\
        \midrule
        Clinc-150 & 0.968 & 0.914 & 0.982 & 0.914 & 0.121 & 0.326 \\
        Banking77 & 0.964 & 0.810 & 0.959 & 0.826 & 0.205 & 0.639 \\
        ATIS & 0.964 & 0.953 & 0.972 & 0.989 & 0.179 & 0.245  \\
        Snips & 0.956 & 0.864 & 0.982 & 0.993 & 0.223 & 0.400 \\
        HWU64 & 0.921 & 0.917 & 0.971 & 0.988 & 0.392 & 0.462\\
        \bottomrule
    \end{tabularx}}
    \caption{Performance comparison between OOS data with the INS data of a BERT model evaluated via softmax confidence scores.}
    \label{performance_table}
\end{table*}

\begin{figure*}[!ht] 
  \centering\scalebox{1.01}{
  \includegraphics[width=\linewidth]{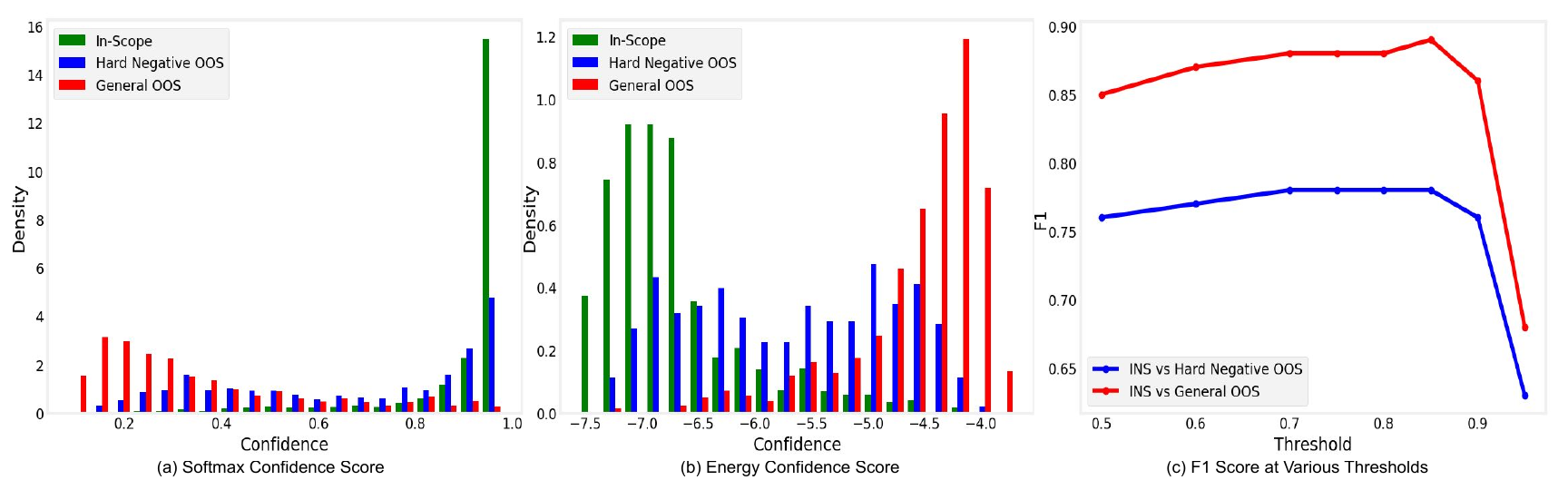} }
  \caption{Results for Banking77 evaluated with BERT. (a) shows the distribution of softmax confidence scores. (b) shows the distribution of energy confidence scores. 
  (c) shows the F1 score of softmax confidence score for hard-negative OOS and general OOS with in-scope at different confidence thresholds. 
  }
  \label{3_sub_figures}
\end{figure*}

\begin{figure}
    \centering\scalebox{0.36}{
    \includegraphics{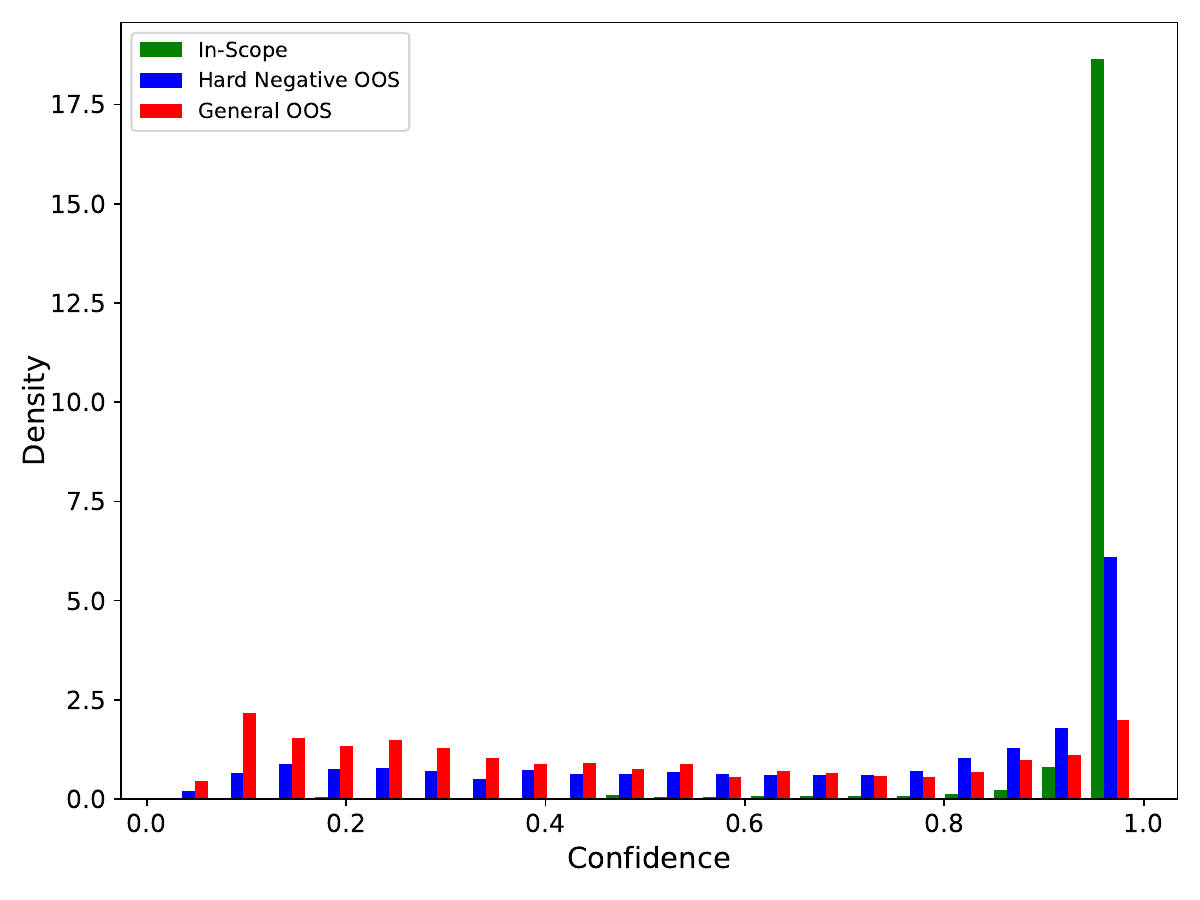}}
    \caption{Distribution of softmax confidence scores for Clinc-150 evaluated with RoBERTa.}
    \label{clinc_conf}
\end{figure}

\begin{table*}
    \centering\scalebox{0.9}{
    \begin{tabularx}{\textwidth}{lXXXXXXXX}
        \toprule
        \multirow{2}{*}{\textbf{Dataset}} & \multicolumn{2}{c}{\textbf{INS}} & \multicolumn{2}{c}{\textbf{General OOS}} & \multicolumn{2}{c}{\textbf{Hard-Negative OOS}} & \multicolumn{2}{c}{\textbf{Both OOS}}\\
        \cline{2-9}
        & \textbf{General OOS} & \textbf{HN OOS} & \textbf{General OOS} & \textbf{HN OOS} & \textbf{General OOS} & \textbf{HN OOS} & \textbf{General OOS} & \textbf{HN OOS}\\
        \midrule
        Clinc-150 & 0.968 & 0.914 & 0.989 & 0.916 & 0.981 & 0.988 & 0.984 & 0.990\\
        Banking77 & 0.964 & 0.810 & 0.997 & 0.874 & 0.989 & 0.996 & 0.996 & 0.992\\
        ATIS & 0.964 & 0.953 & 0.998 & 0.974 & 0.996 & 0.996 & 1.000 & 1.000 \\
        Snips & 0.956 & 0.864 & 0.999 & 0.919 & 0.968 & 0.999 & 0.998 & 0.993\\
        HWU64 & 0.921 & 0.917 & 0.965 & 0.927 & 0.924 & 0.985 & 0.961 & 0.986\\
        \bottomrule
    \end{tabularx}}
    \caption{The AUROC of the OOS data with the INS data of a BERT model. If the AUROC is closer to 1, then the model is less likely to classify an OOS utterance as INS.  We compare how using general OOS, hard-negative OOS, and both OOS datasets in training improves the model's OOS detection capability.}
    \label{OOS_in_training}
\end{table*}

\subsection{Classifier Performance} \label{baseline_evaluation}
We evaluate our five generated hard-negative OOS datasets with BERT, RoBERTa, and DistilBERT trained on only INS data.
Table~\ref{performance_table} shows the AUROC, AUPR, and FPR95 of all datasets evaluated with BERT.

For Banking77 and Clinc-150, the softmax and energy confidence scores for the INS predictions are substantially closer to the hard-negative OOS predictions than the general OOS predictions for all three models. 
% The hard-negative OOS yields a lower AUROC than any of the other examined OOS datasets with the INS data for all three models. For example, when evaluated with BERT, the AUROC of hard-negative OOS with INS for Banking77 is 0.810 while the AUROC of the general OOS with INS is 0.964, indicating that our hard-negative OOS poses a much greater challenge for the model than the general OOS data.
For instance, when evaluated with BERT, Figure~\ref{3_sub_figures} (a) shows that the distribution of the softmax confidence scores for the hard-negative OOS is much closer to the INS data compared to the general OOS data. Figure~\ref{3_sub_figures} (b) demonstrates that the distribution of energy confidence scores for the INS data is also more similar to that of the hard-negative OOS data compared to the general OOS data.
Figure~\ref{3_sub_figures} (c) displays the F1 score for the INS vs hard-negative OOS data is lower than that for the INS vs. general OOS data at all confidence thresholds, indicating the model is worse at distinguishing hard-negative OOS than general OOS from the INS data. 
RoBERTa and DistilBERT resulted in distributions of confidence scores comparable to BERT. Figure~\ref{clinc_conf} displays the distribution of softmax confidence score for RoBERTa after training on the in-scope data from Clinc-150. 

During the evaluation for ATIS, Snips, and HWU64, all three models predict both hard-negative OOS data and general OOS data with a substantial number of high softmax confidence scores.
The AUROC of the softmax and energy confidence scores for our hard-negative OOS with INS are lower than those for general OOS with INS, indicating that our generated hard-negative OOS utterances has a larger overlap with the INS utterances compared to the general OOS and INS's overlap in confidence scores.
The FPR95 for both softmax and energy confidence for hard-negative OOS with INS are higher than those for general OOS with INS across all models and datasets, highlighting that the confidence scores can more effectively differentiate the INS data from the general OOS data than our hard-negative OOS data.

Our results suggest that the hard-negative OOS utterances generated with our approach are, at minimum, as challenging as the general OOS dataset and frequently result in high-confidence, incorrect predictions from intent classifiers.
Specifically for Clinc-150 and Banking77, the hard-negative OOS utterances are substantially more difficult to differentiate as be OOS in comparison to the general OOS utterances.
Therefore, our proposed hard-negative OOS datasets and other hard-negative OOS datasets generated following our approach will  challenge  intent classifiers and scrutinize classifiers' robustness against such data.

\subsection{Using Hard-Negative OOS in training}\label{sec:results-training}
When we train transformer-based intent classifiers only on INS data (Section~\ref{baseline_evaluation}), the models predicted high confidence for hard-negative OOS utterances and varying general OOS confidence across all five benchmark datasets. To determine if using hard-negative OOS in \emph{training} can improve model robustness, we compare the model confidence after training BERT with (1) INS and general OOS data, (2) INS and hard-negative OOS data, (3) INS, general OOS, and hard-negative OOS data, using a 80/20 train test split for the datasets that are used in training.

Table~\ref{OOS_in_training} presents results for the AUROC of the softmax confidence scores to be INS for hard-negative OOS and OOS data with the INS data for each model across five datasets.
Since OOS data are included in training, we consider ``oos'' to be a new label.
We calculate the confidence score for each prediction by taking the highest softmax score for any in-scope intent. 
This confidence score indicates how confident the classifier models are in predicting that utterance belongs to a known intent and is in-scope.

For Clinc-150, Banking77, and ATIS, when the intent classifiers are trained on OOS data, the confidence scores for general OOS utterances drop substantially compared to models trained only on in-scope data, but the confidence score for hard-negative OOS utterances remain high.

In comparison, when we add hard-negative OOS in the training corpora, the models produce low confidence predictions for both hard-negative OOS and OOS utterances, shown by the higher AUROC in Table~\ref{OOS_in_training}. 

% For example, when we train BERT with only the INS data from Banking77, the AUROC for the general OOS and hard-negative OOS with INS were 0.810 and 0.964 respectively. After using general OOS data in training, the AUROC for the general OOS testing data increases to 0.998 but the AUROC for hard-negative OOS data only improves to 0.874. Using our generate hard-negative OOS data in training, the AUROC for OOS and hard-negatives OOS with the INS data changes to 0.989 and 0.996 respectively. 

For Snips and HWU64, incorporating general OOS in training results in high confidence predictions for hard-negative OOS, and using hard-negative OOS in training results in high confidence predictions for the general OOS. 
In this case, incorporating only hard-negative OOS in training in not enough to ensure model robustness. 
When using both hard-negative OOS and general OOS in training, the model predicts OOS utterances with low confidence to be INS.
% Figure~\ref{hard_negative_in_training} shows that when the model is trained on INS and general OOS data, the confidence scores for hard-negative OOS is substantially higher than when hard-negative OOS is used in training.

This result indicates that models trained solely with INS and general OOS data are still prone to predicting hard-negative OOS data with high confidence. When the generated hard-negative OOS datasets are incorporated in the training data, the models are much less likely to produce high confidence scores for hard-negative OOS utterances and displays varying improvements for detecting general OOS utterances. 
Although our experiments demonstrate that training with both hard-negative OOS and general OOS greatly reduces confidence on OOS utterances, we note that there is no way to guarantee that every OOS intent can be covered.
That is, in deployment, we cannot guarantee that the distribution of inputs will follow the OOS data generated with our approach.
Nonetheless, our approach can be used to help improve model robustness and to improve benchmarking and data quality.
%Although in our experiment, training with both hard-negative OOS and general OOS greatly reduce the confidence scores when evaluating all OOS utterances, there is no guarantee that the OOS utterance encountered in deployment will be in the same distribution as the OOS utterances used in training. 
%Therefore, incorporating hard-negative data in training is not guaranteed to lead to a robust intent.
%FIXME but then you're basically invalidating all the work you did.

\section{Conclusion}
We present a new approach to 
%This paper presents a novel approach to 
generating hard-negative OOS data using ChatGPT. After manually reviewing and verifying generated data, we show our approach can generate data that are OOS data with infrequent mislabels. 
% Through evaluating the generate hard-negative OOS datasets with BERT, RoBERTa, and DistilBERT, 
Our evaluation shows that models  trained only on INS data are brittle when tested against hard-negative OOS utterances generated with out approach and often result in overconfident but incorrect predictions. 
% Notably, for datasets such as HWU64 and SNIPS, the model outputs high confidence for both general OOS utterances and hard-negatives utterances. 
Our results indicate that the hard-negative OOS utterances are more challenging to differentiate from the INS utterances when compared to the general OOS utterances.
Furthermore, we show that using hard-negative OOS data in training improves model robustness against hard-negative OOS utterances substantially and general OOS utterances to varying degrees. Models trained on general OOS data still struggle with hard-negative OOS utterances to a noticeable extent across all five datasets. 
Since collecting hard-negative OOS data with ChatGPT is substantially less costly than traditional crowd-sourcing methods, we hope that our technique and analysis will lead to more robust intent classifiers.

All code used for generating and verifying hard-negative OOS data with ChatGPT, datasets generated and used, results, and additional figures are available at \href{https://github.com/frank7li/Generating-Hard-Negative-Out-of-Scope-Data-with-ChatGPT-for-Intent-Classification}{github.com/frank7li/Generating-Hard-Negative-Out-of-Scope-Data-with-ChatGPT-for-Intent-Classification}.

\section*{Acknowledgements}
%We thank Vanderbilt University Summer Research Program (VUSRP) for providing a stipend for this research.
%We acknowledge partial support from the Vanderbilt University Summer Research Program and Seeding Success program.
We acknowledge support from Vanderbilt University's Summer Research and Seeding Success programs.

%\subsection{Appendices}

%Appendices are material that can be read and include lemmas, formulas, proofs, and tables that are not critical to the reading and understanding of the paper, as in \href{https://acl-org.github.io/ACLPUB/formatting.html#appendices}{*ACLPUB}. It is highly recommended that the appendices should come after the references; the main text and appendices should be contained in a `single' manuscript file, without being separately maintained. Letter them in sequence and provide an informative title: \textit{Appendix A. Title of Appendix}

%\subsection{Extra space for ethical considerations and limitations}

%Please note that extra space is allowed after the 8th page (4th page for short papers) for an ethics/broader impact statement and a discussion of limitations. At submission time, if you need extra space for these sections, it should be placed after the conclusion so that it is possible to rapidly check that the rest of the paper still fits in 8 pages (4 pages for short papers). Ethical considerations sections, limitations, acknowledgments, and references do not count against these limits. For camera-ready versions, nine pages of content will be allowed for long (5 for short) papers.

%\nocite{*}
\section*{References}\label{sec:reference}

\bibliographystyle{lrec_natbib}
\bibliography{lrec-coling2024-example}

%\section{Language Resource References}
%\label{lr:ref}

\bibliographystylelanguageresource{lrec_natbib}
\bibliographylanguageresource{languageresource}

%\appendix

%\section{Appendix: How to Produce the \texttt{.pdf}}
%\label{sec:append-how-prod}

%In order to generate a PDF file out of the LaTeX file herein, when citing language resources, the following steps need to be performed:

%\begin{lstlisting}[basicstyle=\ttfamily,frame=single]
%xelatex paper.tex
%bibtex paper.aux
%bibtex languageresource.aux
%xelatex paper.tex
%xelatex paper.tex
%\end{lstlisting}

\end{document}